\documentclass{article}

\usepackage{arxiv}

\usepackage[utf8]{inputenc} 
\usepackage[T1]{fontenc}    
\usepackage{hyperref}       
\usepackage{url}            
\usepackage{booktabs}       
\usepackage{amsfonts}       
\usepackage{amsmath}
\usepackage{nicefrac}       
\usepackage{microtype}      
\usepackage{cleveref}       
\usepackage{graphicx}
\usepackage{natbib}
\usepackage{doi}

\title{Text-conditioned Segmentation for Tomato Phenotyping via Procedural Synthetic Data}

\newif\ifuniqueAffiliation

\ifuniqueAffiliation 
\author{ \href{https://orcid.org/0000-0000-0000-0000}{\includegraphics[scale=0.06]{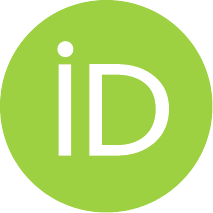}\hspace{1mm}David S.~Hippocampus}\thanks{Use footnote for providing further
		information about author (webpage, alternative
		address)---\emph{not} for acknowledging funding agencies.} \\
	Department of Computer Science\\
	Cranberry-Lemon University\\
	Pittsburgh, PA 15213 \\
	\texttt{hippo@cs.cranberry-lemon.edu} \\
	\And
	\href{https://orcid.org/0000-0000-0000-0000}{\includegraphics[scale=0.06]{orcid.pdf}\hspace{1mm}Elias D.~Striatum} \\
	Department of Electrical Engineering\\
	Mount-Sheikh University\\
	Santa Narimana, Levand \\
	\texttt{stariate@ee.mount-sheikh.edu} \\
}
\else
\usepackage{authblk}

\newbox{\orcid}\sbox{\orcid}{\includegraphics[scale=0.06]{orcid.pdf}} 
\author[1]{%
    \hspace{1mm}Samy Mounir
}
\author[1,2]{%
    \hspace{1mm}Mikolaj Cieslak
}
\author[1]{%
    \hspace{1mm}Najmeddine Dhieb
}
\author[1]{%
    \hspace{1mm}Hakim Ghazzai
    \thanks{\texttt{hakim.ghazzai@kaust.edu.sa}}%
}
\author[1]{%
    \hspace{1mm}Jonathan Klein%
}
\author[1]{%
    \hspace{1mm}Katja Froehlich%
}
\author[4]{%
    \hspace{1mm}Soeren Pirk
}
\author[3]{%
    \hspace{1mm}Wojciech Palubicki
}
\author[5]{%
    \hspace{1mm}Gianluca Setti
}
\author[1]{%
    \hspace{1mm}Ahmed M. Eltawil
}
\author[1]{%
    \hspace{1mm}Dominik L. Michels\thanks{\texttt{dominik.michels@kaust.edu.sa}}
}
\affil[1]{Computer, Electrical and Mathematical Sciences and Engineering Division, KAUST, KSA}
\affil[2]{Department of Computer Science, University of Calgary, Canada}
\affil[3]{Department of Artificial Intelligence, Adam Mickiewicz University, Poland}
\affil[4]{Department of Computer Science, Kiel University, Germany}
\affil[5]{Department of Electronics and Telecommunications, Polytechnic University of Turin, Italy}
\fi

\renewcommand{\shorttitle}{Text-conditioned Segmentation for Tomato Phenotyping}

\hypersetup{
pdftitle={Text-conditioned Segmentation for Tomato Phenotyping via Procedural Synthetic Data},
pdfsubject={},
pdfauthor={},
pdfkeywords={},
}

\begin{document}
\maketitle

\begin{abstract}
    Vision-based automation is an excellent candidate for reducing manual labor in greenhouse crop production and phenotyping. However, progress is constrained by the lack of annotated training data. Recent advances in vision-based foundational models have shown promising results in zero-shot generalization to novel domains, but their performance drops in complex agricultural environments. In this work, we present a sim-to-real framework for tomato plant segmentation that combines synthetic data generation with fine-tuning of a foundation model. We model a commercial cherry tomato greenhouse and use it to generate a large-scale synthetic dataset under diverse viewpoints, lighting conditions, and plant morphology. Subsequently, we fine-tune the Segment Anything Model~3 (SAM~3) on the synthetic dataset, specializing its text-conditioned segmentation behavior for greenhouse crop organs while retaining the general visual prior that makes zero-shot transfer possible. By evaluating our framework on multiple real-world greenhouse datasets, we demonstrate that combining synthetic data with SAM~3 fine-tuning significantly improves segmentation performance and model confidence. To support community benchmarking, we publicly release the procedural model, the generated synthetic dataset, and our fine-tuned SAM~3 weights.
\end{abstract}


\section{Introduction}\label{sec1}

Tomato is an important crop globally, making its phenotyping of great interest to researchers and breeders \cite{Kimura2008,Cobb2013,Rothn2019tde}. However, existing manual phenotyping methods in commercial greenhouses are labor-intensive, costly, and time-consuming \cite{Zhu2022qet,Zhang2026tma}. The need to reduce agricultural labor has driven significant interest in automation and computer-based vision, particularly the deployment of robotic systems and unmanned vehicles for crop monitoring and health assessment \cite{Kasper2020pby, Wang2022phe, Xie2025ph}. Despite this, the complex architectural structure of tomato plants, which is characterized by dense foliage, overlapping leaves, and occluded fruit clusters, makes automated image phenotyping, the fundamental stage in the plant monitoring automation process, challenging. A commercial tomato canopy poses a significant challenge to boundary precision in segmentation. This limits the amount of high-quality data that can be collected and annotated, thereby restricting the performance and scalability of vision-based phenotyping models.

Creating robust vision-based AI for agriculture requires overcoming numerous obstacles, including the lack of large annotated datasets and variation across growing conditions~\cite{White2021dct}. Additionally, collecting and manually annotating such data is expensive, and privacy concerns often limit data sharing among commercial growers. Virtual greenhouses offer a scalable alternative, allowing for the automated generation of photorealistic training data with pixel-perfect ground truth annotations \cite{Cieslak2024gda,Klein2024sds,Khan2025etp}, however, creating them remains a significant challenge. For example, existing 3D models of cherry tomatoes are largely static, unable to simulate continuous growth dynamics or reflect realistic, heavily managed greenhouse conditions, such as pruning and vine lowering. Consequently, there is a lack of publicly available virtual environments suitable for developing and testing vision-based perception systems in realistic greenhouse conditions. Furthermore, even with synthetic data, mismatches in lighting, textures, and plant morphology often lead to poor transfer performance \cite{Cieslak2024gda,Kelly2024winsyn,Lee2024eps}. Finally, while modern vision foundation models like the Segment Anything Model~3 (SAM~3) \cite{Carion2025sam3} exhibit remarkable zero-shot capabilities, their performance degrades when applied to niche agricultural domains \cite{Ji2024snp, Carraro2023sam, Chen2023avfm, Gui2024ees}. Testing these powerful models on agricultural data without sacrificing their generalizability remains an open challenge due to limited training data and domain shift.

In this work, we address these bottlenecks by developing a 3D model of a tomato greenhouse built in Unreal Engine~5. Our approach integrates procedurally generated tomato plants that accurately capture structural variability, growth stages, and greenhouse management practices within a high-fidelity environment. This framework enables the generation of large-scale synthetic datasets under diverse viewpoints and illumination conditions.
Subsequently, we leverage these large synthetic datasets to fine-tune SAM~3 through full fine-tuning and Low-Rank Adaptation (LoRA), followed by weight interpolation with the zero-shot checkpoint. These strategies specialize SAM~3's promptable segmentation behavior for tomato organs while explicitly studying how much of the model's broad pre-trained visual prior should be preserved for real-domain transfer.

The main contributions of this paper are summarized as follows:
\begin{itemize}
    \item We generate a large-scale, photorealistic synthetic dataset with pixel-perfect ground truth annotations for tomato plant phenotyping.
    \item We propose a text-conditioned foundation-model fine-tuning pipeline for greenhouse crop phenotyping, fine-tuning SAM~3 via synthetic parameter-efficient fine-tuning, full fine-tuning, and weight interpolation.
    \item We evaluate our approach on three real-world tomato segmentation datasets, demonstrating that our best SAM~3 variant improves zero-shot macro fruit Intersection over Union (IoU) by 12.9 percentage points (from 52.7\% to 65.6\%).
    \item We publicly release on Hugging Face the complete synthetic dataset (RGB, depth, semantic, and instance masks), the fine-tuned SAM~3 weights, and updated annotations for public datasets to facilitate benchmarking and future research.
\end{itemize}

\section{Related Work}\label{sec2}

\paragraph{Vision foundation models for agriculture:} The need to reduce annotation labor has driven interest in large-scale vision models, yet their application to agriculture remains limited by significant domain shifts from natural images. Recently, several works have advanced agricultural vision by applying Vision Foundation Models (VFMs) to downstream tasks. Al Nahian et al. \cite{AlNahian2025AgriFM} proposed Agri-FM+, a self-supervised foundation model for agricultural vision trained via a two-stage continual learning pipeline. Singh et al. \cite{Singh2025FewShotAO} applied the grounding DINO model to agricultural object detection, developing a few-shot method that eliminates reliance on text prompts. Lundqvist et al. \cite{Lundqvist2026dysp} used prompt optimization to evaluate open-vocabulary detectors (including SAM~3) for cowpea flower and pod detection, showing that prompt engineering can close the gap between zero-shot VFMs and supervised detectors, and that optimal prompts are model-specific, non-obvious, and transferable. Shen et al. \cite{Shen2026afpv} combined Depth Anything and SAM~3 for accurate canopy segmentation with high counting accuracy and plant ordering. While these methods advance foundation models in agriculture, phenotyping often requires finer organ boundaries and structural granularities than broad pre-training emphasizes. Models like SAM~3 provide a strong generalization basis, but their zero-shot performance can remain poor under dense canopies, heavy occlusion, fine organ boundaries, and crop-specific taxonomies. Full fine-tuning, Parameter-Efficient Fine-Tuning (PEFT) \cite{Xu2026PEFT}, or weight interpolation can specialize that basis without training from scratch. Chen et al. \cite{Chen2023avfm} showed that PEFT, specifically LoRA, efficiently adapts models like DINOv2 to plant phenotyping tasks such as leaf counting and disease classification on real data. However, fine-tuning VFMs for promptable segmentation (like SAM~3) under strict sim-to-real domain shifts remains unexplored.

\paragraph{Synthetic image generation for agriculture:} The lack of large-scale annotated datasets in precision agriculture has driven the adoption of synthetic image generation for training deep learning models. Early work focused on model organisms such as \textit{Arabidopsis} \cite{Ubbens2018upm,Ward2018deep}, and later research expanded to aerial imagery \cite{Chiu2020AgriVision}, simplified outdoor environments \cite{Olsen2019deepweeds}, and indoor farming scenes \cite{Anagnostopoulou2023synthetic,Zheng2019CropDeep}. Datasets can be built using 2D image composition \cite{Gao2020dcn} or static 3D asset libraries \cite{Angarano2023domain,Hu2022mra}. Hartely et al. \cite{Hartley2025plantdreamer} proposed a diffusion-guided Gaussian splatting pipeline for realistic herbaceous 3D plant models. To bridge the domain gap, others employ generative and domain adaptation techniques: Fei et al. \cite{Fei2021synthetic2real} used GANs to translate 3D grapevine renders into photorealistic images, and Sapkota et al. \cite{Sapkota2022usi} generated fake weed instances for augmentation in cotton fields. More recently, Heschl et al. \cite{Heschl2025synthset} introduced SynthSet, a dual diffusion architecture that synthesizes agricultural imagery with pixel-perfect annotations without human intervention. Procedural generation with rendering engines and L-systems offers greater structural control for complex canopies, notably in wheat head segmentation \cite{Beheshtifard2025,Napier2023swl}, further refined by semi-supervised and generative pipelines to handle domain shifts \cite{Myers2024eths,Najafian2023sls}. Beyond wheat, procedural modeling spans multiple species: Barth et al. \cite{Barth2018dsm} rendered sweet pepper meshes from empirical measurements, Cieslak et al. \cite{Cieslak2024gda} built soybean datasets, and Khan et al. \cite{Khan2025etp} evaluated L-system data for maize and canola, showing that procedural calibration to real images can improve accuracy ten-fold. Wang et al. \cite{Wang2025dsi} leveraged synthetic data for semantic and instance segmentation in fruit orchards. Despite these advances, comprehensive synthetic datasets for complex tomato greenhouses remain scarce. Klein et al. \cite{Klein2024sds} generated a synthetic tomato dataset limited to disease detection. We address this gap with a procedural pipeline tailored to commercial tomato greenhouses and a large-scale, multimodal dataset for complex phenotyping.

\paragraph{Tomato plant modeling:} Plant models are widely used to simulate the growth and architectural development of tomatoes \cite{Sarlikioti2011hpa, Chen2014qea, DeVisser2014oii, Zhang2020hrs, Butturini2025afs, Deng2024gps, Filippo2023api}, focusing either on physiological processes, as in Functional Structural Plant Models (FSPMs), or on geometric representations of plant architecture. A detailed discussion of the biological basis of our L-system model is provided in Appendix \ref{appendix:model_bio_basis} and the related literature in Appendix \ref{appendix:tomato_plant_modeling}.

\section{Materials and Methods}\label{sec3}

This section details the methodologies and datasets employed in our framework. We begin by outlining the real-world data used for evaluation, comprising both established public datasets and a novel image collection captured within a commercial greenhouse. Next, we present the architecture of our photorealistic virtual greenhouse, describing the procedural generation of the tomato plant models and their integration into the Unreal Engine~5 greenhouse environment. Finally, we discuss the generation of the large-scale synthetic dataset. 

\subsection{Collection of Existing Public Datasets}
We drew on public tomato datasets that span both mask- and bounding-box-level annotations. Three datasets provide segmentation masks suitable for semantic evaluation and serve as the basis for our quantitative real-domain evaluation: TomatoMAP-Seg \cite{Zhang2026tma}, LaboroTomato \cite{trigubenko2020laborotomato}, and Rob2Pheno \cite{Manya2020tfd}. Three additional bounding-box datasets TomatoOD \cite{tsironis2020tomatod}, AgRobTomato \cite{Moreira2022bdl}, and Tomato Plantfactory \cite{Wu2023tpf} are included in Table~\ref{tab:dataset_coverage} for completeness and are part of the released benchmark suite. A full detection evaluation on these is left to future work (see Section~\ref{sec5}). We explicitly restricted our focus to tomato fruit detection and segmentation and excluded other publicly available datasets that target unrelated annotation formats or target classes. Table~\ref{tab:dataset_coverage} summarizes the dataset coverage.

\begin{table}[t]
    \centering
    \caption{Summary of datasets. The three mask datasets are used for semantic fruit IoU evaluation in this work. The three box datasets are listed for completeness as part of the released benchmark suite, with detection evaluation left to future work.}
    \label{tab:dataset_coverage}
    \resizebox{\columnwidth}{!}{%
        \begin{tabular}{lcccl}
            \toprule
            \textbf{Dataset}    & \textbf{Images} & \textbf{Instances} & \textbf{Segmentation Masks} & \textbf{Main Use}                   \\
            \midrule
            TomatoMAP-Seg       & 725             & 4,196              & Yes                         & Semantic fruit (evaluated)          \\
            LaboroTomato        & 804             & 9,777              & Yes                         & Semantic fruit (evaluated)          \\
            Rob2Pheno           & 105             & 1,612              & Yes                         & Semantic fruit, non-exhaustive labels \\
            AgRobTomato         & 435             & 6,084              & No                          & Box (future work)                   \\
            TomatoOD            & 277             & 2,421              & No                          & Box (future work)                   \\
            Tomato Plantfactory & 520             & 9,112              & No                          & Box (future work)                   \\
            \bottomrule
        \end{tabular}%
    }
\end{table}

\subsection{Collection and Processing of Images for the Real Dataset}

To evaluate our models under operational conditions, we captured a real-world dataset in a commercial tomato greenhouse in Saudi Arabia. Data were acquired using a 20\,MP RGB camera. The recordings captured natural variations in plant geometry and occlusion from various camera angles and viewpoints.
As this dataset lacks ground-truth annotations, we used it solely for qualitative evaluation of our models (e.g., comparing zero-shot SAM~3 with our fine-tuned variants).
Because of privacy concerns and the proprietary nature of the commercial greenhouse, we do not disclose its exact details.

\subsection{Synthetic Dataset Generation}

While existing synthetic datasets for complex tomato greenhouse environments often rely on static assets or simplify plant architecture, accurately simulating these facilities requires modeling both biological growth and intensive human intervention. To address this, we developed a novel procedural model of cherry tomato plants (\textit{Solanum lycopersicum} var.\ cerasiform) using L-systems \cite{Prusinkiewicz1990tab,Mech1996vmp,Cieslak2021lmi}. Our model captures sympodial branching patterns, and vegetative and reproductive development according to established botanical rules \cite{Went1944tomato, TorresQuezada2023tomato}. The primary contribution of our model lies in its ability to simulate commercial greenhouse management practices. By incorporating biomechanical tropisms for natural posture alongside algorithmic pruning and the leaning-and-lowering technique, the model generates canopies that reflect operational conditions. Stochastic components modulate geometric parameters to ensure that no two simulated plants are morphologically identical. Detailed descriptions of the biological basis, L+C language implementation, tropism simulation, and stochastic parameters are provided in Appendix \ref{appendix:l_system_model}.

\begin{figure}[t]
    \centering
    \includegraphics[width=\textwidth, height=0.37\textheight]{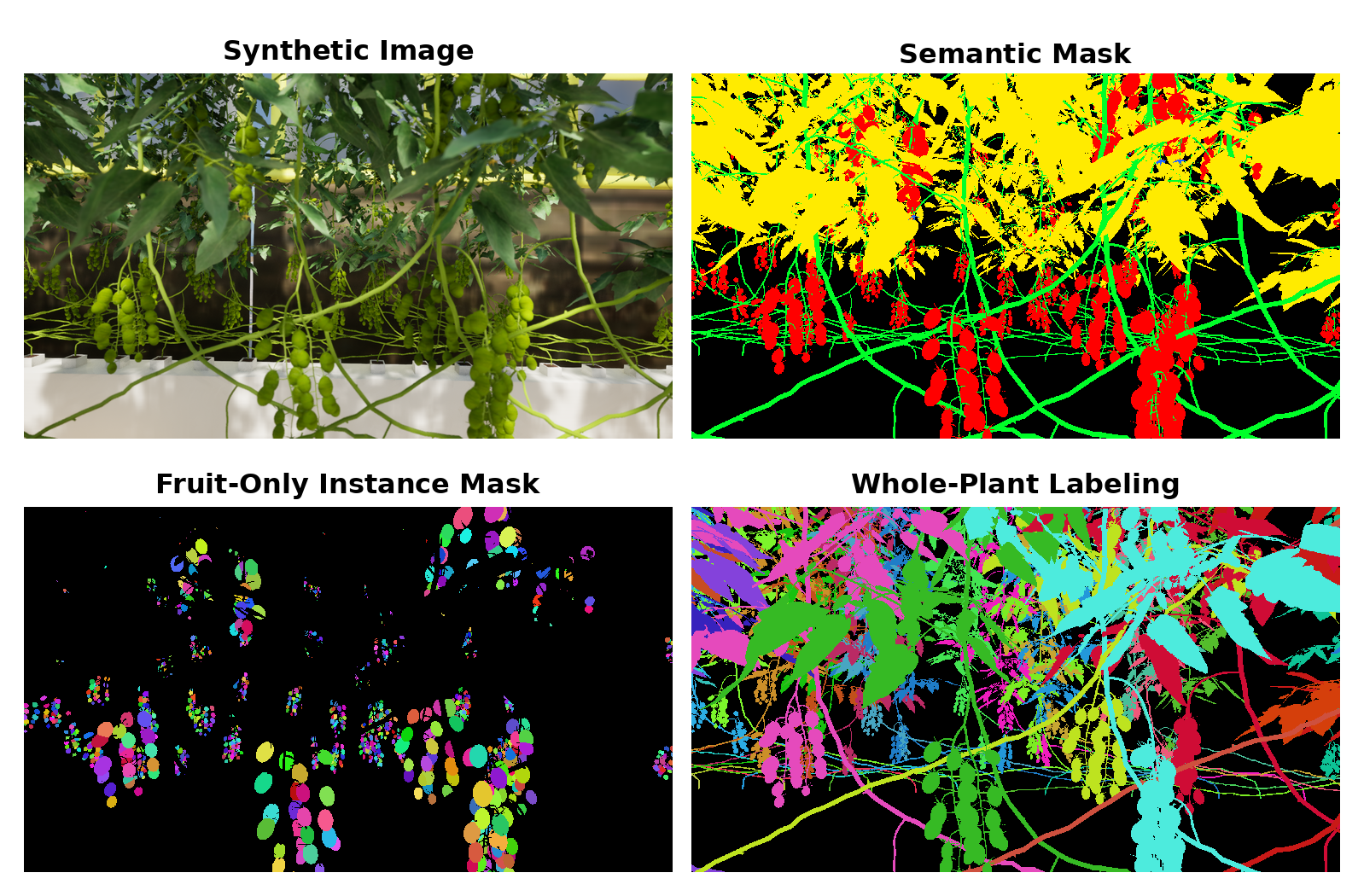}
    \caption{Example of a synthetic image and its corresponding labels. Semantic masks are colored red for fruit, green for stems, yellow for leaflets, and blue for flowers. Instance masks are shown for individual fruits and whole plants.}
    \label{fig:synthetic-example}
\end{figure}

Leveraging this procedural model, we generated a large-scale synthetic dataset for training machine learning models. We implemented a configurable image-acquisition pipeline that supports multiple virtual camera placements within the Unreal Engine~5 environment. For each capture instance, the simulator generates synchronized RGB images, depth maps, and pixel-perfect annotations, including instance masks, semantic labels (leaf, stem, flower, fruit, background), and bounding boxes. Figure~\ref{fig:synthetic-example} shows an example of a synthetic image and its corresponding labels. More details on the metadata structure and the image acquisition pipeline are provided in Appendix \ref{appendix:synthetic_dataset}.

\subsection{Fine-Tuning Foundation Models (SAM 3)}
Recent advancements in vision foundation models, such as SAM~3 \cite{Carion2025sam3}, offer powerful zero-shot segmentation capabilities through interactive prompts (points, bounding boxes, or text). However, their performance often degrades when applied to complex, dense agricultural scenes without domain-specific training \cite{Carraro2023sam, Ji2024snp, Gui2024ees, AlNahian2025AgriFM}. To fine-tune SAM~3 for the greenhouse domain while preserving its generalizable features, we explored three complementary training and weight-interpolation strategies using our synthetic dataset. The full fine-tuning and LoRA runs used the four paper target classes (leaf, stem, flower, and fruit), and the final real-domain semantic inference used multi-phrase prompt groups for those same four organs:

\paragraph{Parameter-Efficient Fine-Tuning:} We applied Low-Rank Adaptation (LoRA) \cite{Hu2022lora} on the attention layers of the SAM~3 vision encoder. PEFT methods update only a small fraction of the model parameters, preserving the majority of the pre-trained representations. We trained LoRA on the 68k synthetic corpus for 5 epochs. The optimizer was restricted to LoRA adapter tensors on the last 12 vision-transformer attention qkv layers; the base model, decoder, semantic head, and other heads were not optimizer targets.

\paragraph{Full Fine-Tuning:} We trained all semantic components of SAM~3, including the semantic segmentation head, on the 68k synthetic corpus with a 10-epoch cosine learning rate schedule and sliding-window evaluation. This provides a strong domain-specialized baseline that modifies the full parameter space.

\paragraph{Weight Interpolation:} To preserve the zero-shot generalization capabilities of the foundation model while incorporating domain-specific knowledge, we applied weight-space interpolation between the pre-trained SAM~3 checkpoint and a fine-tuned checkpoint. Following WiSE-FT \cite{Wortsman2021wiseft}, we computed the interpolated weights as $\theta_{\text{wise}}(\alpha) = (1-\alpha)\cdot\theta_{\text{base}} + \alpha\cdot\theta_{\text{ft}}$ with $\alpha=0.75$, giving a 25\% zero-shot contribution and a 75\% fine-tuned-checkpoint contribution. For the full fine-tuned model, this interpolation was applied to the full matching floating-point state. For LoRA, we first merged the adapter update into the base checkpoint, because the merged LoRA checkpoint differs from the base only in the LoRA-targeted attention qkv matrices of the last 12 vision-transformer blocks, WiSE-FT LoRA changes only that subset and leaves the remaining tensors at their zero-shot values. This approach draws on WiSE-FT, which demonstrates that linear weight interpolation yields robust out-of-distribution generalization, and \textit{model soups} \cite{Wortsman2022modelsoups}, which show that averaging weights of differently fine-tuned models can improve accuracy without additional inference cost.

\section{Results}\label{sec4}

We evaluated eight model variants across synthetic and real-world datasets to assess the effectiveness of different fine-tuning strategies for agricultural semantic segmentation. The evaluation spanned the three real-world mask datasets from Table \ref{tab:dataset_coverage}, our custom greenhouse dataset, and the synthetic validation set.

The primary objective was to determine the extent to which fine-tuning SAM~3 on synthetic data improves its generalization to real greenhouse imagery, and which fine-tuning strategy yields the best sim-to-real transfer. To evaluate generalization in a task-relevant manner, we focused on semantic segmentation, which assigns each pixel to an organ category (leaf, stem, flower, fruit, background).

We established an evaluation framework comparing eight models. These include the Zero-shot SAM~3 baseline and four fine-tuned variants trained on our 68k synthetic corpus: LoRA SAM~3, Full FT SAM~3, and their weight-interpolated counterparts (WiSE-FT SAM~3 and WiSE-FT LoRA SAM~3). We also evaluated three supervised baselines trained from scratch on the exact same synthetic dataset: DeepLabV3+ \cite{Chen2018DeepLabv3Plus}, SegFormer \cite{Xie2021segformer}, and the domain-specific EoMT \cite{Kerssies2025eomt}.

\subsection{Quantitative Evaluation of the Synthetic Dataset}

We first evaluate the main model variants on our procedurally generated synthetic dataset using a 5,553-frame common-clean validation set with sliding-window inference. Table~\ref{tab:synthetic_eval} summarizes the four-class semantic segmentation results. Semantic segmentation performance is evaluated using per-class mean foreground IoU (mFG) and mean IoU with background (mIoU w/ bg). The supervised baselines achieved the highest synthetic scores (DeepLabV3+: 0.6953 mFG). The full fine-tuned model reached 0.6158 mFG, while WiSE-FT reached 0.6085 mFG. LoRA scored lower on synthetic validation (0.5502 mFG, 0.5965 mIoU w/ bg) but transferred substantially better to real imagery than its synthetic ranking alone would suggest.
This result suggests that supervised architectures overfit to the synthetic domain. Conversely, SAM~3 is constrained by strong pre-trained priors, which prevents it from memorizing the synthetic data and lowers its scores. However, as demonstrated in the next section, this is what enables SAM~3 to achieve better generalization on real-world data.

\begin{table}[t]
    \centering
    \caption{Four-class semantic segmentation on the synthetic validation set (5,553 frames, sliding-window inference). Models are ordered by mean foreground IoU (mFG). Synthetic validation alone does not predict real-world transfer (see Table~\ref{tab:semantic_seg}).}
    \label{tab:synthetic_eval}
    \begin{tabular}{llcc}
        \toprule
        \textbf{Model}  & \textbf{Family} & \textbf{mFG}    & \textbf{mIoU w/ bg} \\
        \midrule
        DeepLabV3+      & Supervised      & \textbf{0.6953} & \textbf{0.7406}     \\
        EoMT         & Supervised      & 0.6773          & 0.6553              \\
        SegFormer       & Supervised      & 0.6178          & 0.6750              \\
        Full FT SAM~3   & Full FT         & 0.6158          & 0.6678              \\
        WiSE-FT SAM~3   & Weight mix      & 0.6085          & 0.6616              \\
        WiSE LoRA SAM~3 & Weight mix      & 0.5582          & 0.6168              \\
		LoRA SAM~3      & PEFT            & 0.5502         & 0.5965            \\
        Zero-shot SAM~3 & Baseline        & 0.3014          & 0.3680              \\
        \bottomrule
    \end{tabular}
\end{table}

\subsection{Quantitative Evaluation on External Real-World Datasets}

We evaluate all model variants on external real-world datasets to assess cross-domain generalization. Semantic segmentation predictions are reduced to binary fruit-vs-non-fruit to enable direct comparison across datasets with different annotation granularity. We compute fruit IoU as $\text{TP} / (\text{TP} + \text{FP} + \text{FN})$, macro fruit IoU as the unweighted mean across datasets, and image-weighted fruit IoU weighted by dataset size.

An important caveat about labels applies: LaboroTomato and Rob2Pheno are fruit-mask datasets with non-exhaustive annotations. On Laboro, models can be penalized for segmenting additional visible fruits absent from the ground truth. On Rob2Pheno, many visible fruits are unlabeled, which can inflate false-positive counts for comprehensive models. These caveats are most pronounced on Rob2Pheno, where standard IoU can underestimate qualitative segmentation quality.

The quantitative results for semantic segmentation are summarized in Table~\ref{tab:semantic_seg}.
Full FT SAM~3 achieved the best macro fruit IoU (0.6562) and image-weighted fruit IoU (0.7372) across the three mask datasets. LoRA and WiSE-FT LoRA also transferred strongly to real imagery, reaching macro fruit IoU of 0.6099 and 0.6118 respectively; WiSE-FT LoRA achieved the best LaboroTomato fruit IoU (0.7728). WiSE-FT SAM~3 improved over the zero-shot baseline on LaboroTomato and Rob2Pheno but not TomatoMAP-Seg, yielding 0.5984 macro fruit IoU. The supervised baselines (DeepLabV3+, SegFormer, EoMT) achieved non-zero transfer on unseen real data but remained far below the fine-tuned SAM~3 variants, with macro fruit IoU below 0.23.

\begin{table}[t]
    \centering
    \caption{External Semantic Fruit IoU across Real-World Mask Datasets. Predictions are reduced to binary fruit vs. non-fruit. Best per-column values in bold.}
    \label{tab:semantic_seg}
    \resizebox{\columnwidth}{!}{%
        \begin{tabular}{llccccc}
            \toprule
            \textbf{Model}     & \textbf{Family} & \textbf{TomatoMAP} & \textbf{Laboro} & \textbf{Rob2Pheno} & \textbf{Macro}  & \textbf{Img-Wt.} \\
            \midrule
            Full FT SAM~3      & Full FT         & \textbf{0.7785}    & 0.7371          & \textbf{0.4529}    & \textbf{0.6562} & \textbf{0.7372}  \\
            WiSE-FT LoRA SAM~3 & Weight mix      & 0.6407             & \textbf{0.7728} & 0.4219             & 0.6118          & 0.6917           \\
            LoRA SAM~3         & PEFT            & 0.6821             & 0.7234          & 0.4243             & 0.6099          & 0.6859           \\
            WiSE-FT SAM~3      & Weight mix      & 0.6192             & 0.7382          & 0.4378             & 0.5984          & 0.6661           \\
            Zero-shot SAM~3    & Baseline        & 0.5938             & 0.6875          & 0.2995             & 0.5269          & 0.6210           \\
            SegFormer          & Supervised      & 0.1019             & 0.1391          & 0.4381             & 0.2264          & 0.1418           \\
            DeepLabV3+         & Supervised      & 0.0339             & 0.0850          & 0.0020             & 0.0403          & 0.0570           \\
            EoMT            & Supervised      & 0.0065             & 0.0043          & 0.0340             & 0.0149          & 0.0072           \\
            \bottomrule
        \end{tabular}%
    }
\end{table}

\subsection{Qualitative and Label-Free Evaluation on Real Greenhouse Imagery}

Since the real-world greenhouse dataset lacks ground-truth annotations, we employed two complementary strategies: visual overlay comparison and label-free confidence metrics. While these metrics quantify certain aspects of model behavior on real imagery, they are purely indicative. This evaluation is intended to demonstrate the model's viability and consistency in a realistic greenhouse environment.

For each image, SAM~3 was queried with the semantic prompts \textit{leaf}, \textit{stem}, \textit{flower}, and \textit{fruit}. The semantic head output one logit map per prompt, which was converted to probabilities via the sigmoid function. We defined the maximum foreground probability per pixel as $\max(p_{\text{leaf}}, p_{\text{stem}}, p_{\text{flower}}, p_{\text{fruit}})$. A pixel was classified as foreground when the maximum exceeded 0.5. Otherwise, it was assigned to the background. We reported all-pixel confidence (dataset mean of maximum foreground probability), Test-Time Augmentation (TTA) agreement (consistency across identity, horizontal flip, and JPEG compression augmentations), tile disagreement (label conflicts across overlapping sliding-window tiles), and foreground fraction (proportion of pixels predicted as foreground).

\begin{figure}[t]
    \centering
    \includegraphics[width=0.8\textwidth]{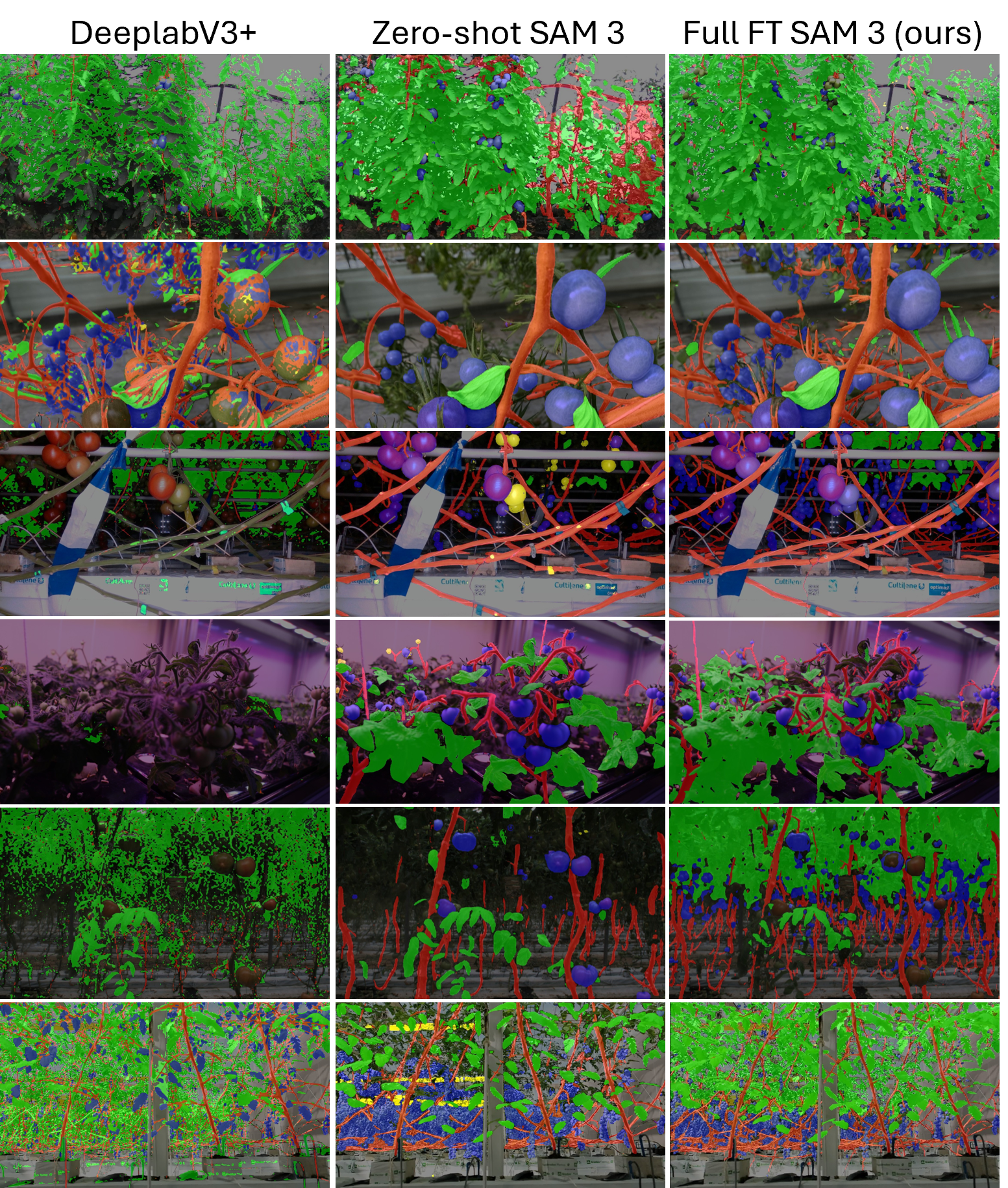}
	\caption{Qualitative comparison of the supervised model DeepLabV3+, zero-shot SAM3, and our fully fine-tuned SAM3 (Full FT SAM3) across six datasets: AgRobTomato, LaboroTomato, Rob2Pheno, Tomato Plantfactory, TomatoOD, and our real greenhouse dataset. The prediction panels show semantic segmentation outputs for leaves (green), stems (red), flowers (yellow), and fruits (blue) overlaid on the original RGB images.}
    \label{fig:real-example}
\end{figure}

Table~\ref{tab:greenhouse_confidence} compares WiSE-FT SAM~3 against the zero-shot baseline on the greenhouse dataset. WiSE-FT achieves substantially higher confidence (+0.24 absolute), won on 350 of 370 matched frames, and demonstrated superior perturbation stability (TTA agreement: 0.9045 vs. 0.7345). The zero-shot model exhibited systematic failure modes: foreground under-coverage (foreground fraction 0.3385 vs. 0.6574), canopy collapse into background, and flower hallucination (flower fraction 0.0471 vs. 0.0005). Visual review, shown in Figure~\ref{fig:real-example}, confirms these label-free findings: fine-tuned SAM~3 shows more coherent labels across different greenhouse images.

\begin{table}[b]
    \centering
    \caption{Label-free confidence metrics on the greenhouse dataset. WiSE-FT SAM~3 achieves higher confidence, better TTA stability, and more realistic foreground coverage than the zero-shot baseline.}
    \label{tab:greenhouse_confidence}
    \begin{tabular}{lccc}
        \toprule
        \textbf{Metric}          & \textbf{WiSE-FT} & \textbf{Zero-shot} & \textbf{Reading}        \\
        \midrule
        All-pixel max confidence & 0.6114           & 0.3685             & WiSE-FT ($\uparrow 0.24 $)         \\
        Frames with higher conf. & 350/370          & 20/370             & WiSE-FT dominant        \\
        Mean entropy             & 0.1374           & 0.1374             & Entropy alone tied      \\
        Foreground TTA agreement & 0.9045           & 0.7345             & WiSE-FT more stable     \\
        Tile disagreement        & 0.0952           & 0.1415             & WiSE-FT fewer conflicts \\
        Foreground fraction      & 0.6574           & 0.3385             & Zero-shot under-covers  \\
        Leaf fraction            & 0.4252           & 0.1331             & Zero-shot misses canopy \\
        \bottomrule
    \end{tabular}
\end{table}

Table~\ref{tab:greenhouse_compact} presents compact confidence metrics across the verified WiSE-FT and full fine-tuning variants.
We define the six reported metrics as follows.
Let $p_i = \max(p_{\text{leaf}},p_{\text{stem}},p_{\text{flower}},p_{\text{fruit}})$ be the maximum foreground sigmoid probability at pixel~$i$, and let $\mathcal{F}=\{i : p_i \ge 0.5\}$ and $\mathcal{B}=\{i : p_i < 0.5\}$ denote the foreground and background pixel sets, respectively.
All-pixel confidence (All-px conf) is the dataset mean of $p_i$ over all pixels.
Foreground confidence (FG conf) is the mean of $p_i$ restricted to $\mathcal{F}$; analogously, background confidence (BG conf) is the mean of $1-p_i$ over $\mathcal{B}$, measuring certainty of background assignment.
Entropy is the mean binary entropy $-p_c \log p_c - (1-p_c) \log(1-p_c)$ averaged across all four organ probability maps.
TTA probability variance (TTA var) is the mean per-pixel variance of individual class probabilities $p_c$ across three test-time augmentations (identity, horizontal flip, and JPEG compression), where lower values indicate more stable predictions.
Background fraction (BG frac) is $|\mathcal{B}|/(|\mathcal{F}|+|\mathcal{B}|)$, the proportion of pixels classified as background.
WiSE-FT achieves the highest all-pixel confidence and foreground confidence, with the lowest TTA probability variance, confirming its robustness on unlabeled greenhouse imagery.

\begin{table}[t]
    \centering
    \caption{Greenhouse compact confidence metrics. Higher confidence and lower TTA variance indicate more robust predictions.}
    \label{tab:greenhouse_compact}
    \footnotesize
    \begin{tabular}{lcccccc}
        \toprule
        \textbf{Model} & \textbf{All-px conf} & \textbf{FG conf} & \textbf{BG conf} & \textbf{Entropy} & \textbf{TTA var} & \textbf{BG frac} \\
        \midrule
        WiSE-FT        & 0.6114               & 0.8829           & 0.8969           & 0.1374           & 0.0022           & 0.3416           \\
        Full FT        & 0.5929               & 0.8686           & 0.8782           & 0.1439           & 0.0039           & 0.3620           \\
        \bottomrule
    \end{tabular}
\end{table}

\begin{figure}[]
    \centering
    \includegraphics[width=\textwidth]{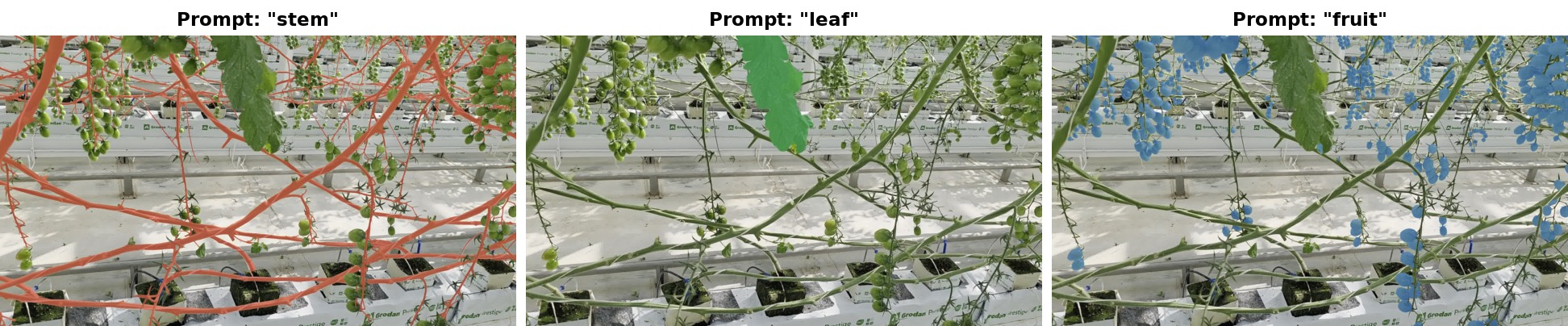}
    \caption{Prompt simplicity analysis. A single image is segmented using concise organ-level prompts: ``stem'', ``leaf'' and ``fruit''. The fine-tuned SAM~3 model cleanly isolates the respective anatomical structures without ambiguity, demonstrating strong alignment with basic taxonomic vocabulary.}
    \label{fig:prompt_simplicity}
\end{figure}

A key advantage of fine-tuning SAM~3 is that it preserves a practical text interface while improving dense agricultural segmentation. As illustrated in Figure~\ref{fig:prompt_simplicity}, concise organ-level prompts such as ``stem'', ``leaf'', and ``fruit'' yield precise, non-overlapping semantic masks from the fine-tuned SAM~3 model. This confirms that synthetic fine-tuning can specialize SAM~3 for heavily intertwined anatomical parts without discarding its prompt-conditioned behavior.

\section{Discussion}\label{sec5}
The evaluation results disclose critical insights into the relationship between synthetic-domain performance and real-world generalization during the fine-tuning of foundation models for agricultural vision tasks.

\subsection{Key Takeaways and Benefits}

\paragraph{Generalization via Synthetic Data:} The all-real semantic audit confirms that our procedurally generated synthetic dataset can bridge the sim-to-real gap when paired with SAM~3 fine-tuning. Full FT SAM~3 achieves the strongest aggregate transfer, with 0.6562 macro fruit IoU and 0.7372 image-weighted fruit IoU across the three external mask datasets. LoRA and WiSE-FT LoRA also transfer well, reaching 0.6099 and 0.6118 macro fruit IoU respectively, compared with 0.5269 for zero-shot SAM~3. These gains do not come from learning agricultural perception from scratch: SAM~3 already provides a broad visual and promptable-segmentation prior. Synthetic fine-tuning specializes that prior for dense tomato canopies, helping the model learn where a fruit ends and an intersecting stem begins, resulting in tighter masks and more complete foreground coverage, addressing known limitations of zero-shot models in accurately segmenting dense canopies \cite{Shen2026afpv}.

\paragraph{Label-Free Confidence on Real Greenhouse Imagery:} On the unlabeled greenhouse dataset, the label-free confidence analysis provides a complementary evaluation pathway. WiSE-FT SAM~3 achieves higher all-pixel confidence (0.6114 vs. 0.3685 for zero-shot), wins on 350 of 370 matched frames, and demonstrates superior TTA agreement (0.9045 vs. 0.7345) and lower tile disagreement. Critically, the zero-shot model exhibits systematic failure modes including foreground under-coverage, canopy collapse into background, and flower hallucination \cite{Ji2024snp, Carraro2023sam, Gui2024ees}, which are captured by the confidence gap and foreground fraction metrics. These label-free proxies are validated by visual review, confirming that fine-tuned SAM~3 variants produce more coherent greenhouse overlays than the zero-shot baseline.

\paragraph{Prompt Simplicity:} A key advantage of our approach is the reliance on straightforward organ-level text prompts at evaluation time. Ultimate-users, such as agronomists or operators of automated harvest systems, require systems that respond reliably to standardized botanical terms without complex prompt engineering \cite{Singh2025FewShotAO}, as finding optimal prompts for agricultural tasks is often model-specific and non-obvious \cite{Lundqvist2026dysp}. Our results demonstrate that fine-tuned SAM~3 models can cleanly segment highly complex canopies from concise prompt groups centered on ``stem'', ``leaf'', and ``fruit''.

\paragraph{Weight Interpolation as Regularization:} WiSE-FT interpolates the fine-tuned checkpoint with the pre-trained SAM~3 base checkpoint using $\theta_{\text{wise}}(\alpha) = (1-\alpha)\theta_{\text{base}} + \alpha\theta_{\text{ft}}$ and $\alpha=0.75$, so 25\% of the weight value comes from the zero-shot checkpoint. For Full FT this is a model-wide interpolation; for LoRA it is restricted to the merged LoRA update tensors, with the rest of the model remaining at the zero-shot weights. This weight-space averaging, inspired by WiSE-FT~\cite{Wortsman2021wiseft} and model soups~\cite{Wortsman2022modelsoups}, serves as implicit regularization, preserving general visual representations while incorporating domain-specific features learned from synthetic tomato data. Consequently, interpolation does not uniformly dominate full fine-tuning: Full FT is strongest on aggregate, while WiSE-FT LoRA achieves the best LaboroTomato fruit IoU (0.7728). This indicates that the benefits of interpolation depend on the fine-tuned checkpoint or LoRA-updated subset and mixture coefficient.

\subsection{Limitations and Evaluation Challenges}

\paragraph{The Synthetic--Real Inversion:} A finding of this work is the mismatch between synthetic and real-domain performance across fine-tuning and weight-interpolation strategies. On the synthetic validation set, supervised baselines and full fine-tuning rank above LoRA, while WiSE-FT trails full fine-tuning by a slight margin. On real-world external datasets, however, LoRA transfers far better than its synthetic ranking would suggest. The supervised baselines are clearly specialized to the synthetic greenhouse domain and produce meaningful non-zero performance on unseen real datasets, but they lack the broad visual prior of SAM~3 and remain far below the SAM~3 variants. This synthetic--real inversion suggests that optimizing for synthetic metrics alone is insufficient and that real-domain evaluation is essential for model selection. It also highlights the central advantage of SAM~3 fine-tuning: the synthetic data does not need to create general visual understanding from scratch, but instead steers an already general model toward the crop-specific structures needed for tomato phenotyping.

\paragraph{PEFT Sensitivity.} Our LoRA results show that PEFT can transfer to real data when the trainable surface is restricted, aligning with observations in real-to-real phenotyping \cite{Chen2023avfm}. LoRA reaches fruit IoUs of 0.6821 (TomatoMAP-Seg), 0.7234 (LaboroTomato), and 0.4243 (Rob2Pheno). WiSE-FT LoRA further improves LaboroTomato fruit IoU to 0.7728. In sim-to-real settings, PEFT capacity, update placement, and interpolation strength are critical: overly broad low-rank updates can overfit synthetic-domain statistics, while focused attention-layer updates preserve enough pre-trained structure to transfer. 

\newpage

\paragraph{Evaluation Rigidities.} A primary limitation is the rigid evaluation framework inherited from standard object detection metrics, which struggles to accommodate the hierarchical nature of plant morphology \cite{roggiolani2022hierarchical, Xing2026, Kelly2016}. The fruit IoU metric, while informative, does not capture full four-class semantic quality due to the absence of exhaustive multi-organ annotations in existing real-world datasets.

\paragraph{Taxonomy Granularity and Annotation Caveats:} A significant challenge in evaluation arises from annotation quality and taxonomy mismatches across datasets. Rob2Pheno contains non-exhaustive fruit labels, where many visible fruits are unlabeled, inflating false-positive counts for models that segment comprehensively. On LaboroTomato, WiSE-FT SAM~3 appears slightly penalized for segmenting additional visible fruits that are absent from the ground truth. These caveats highlight that standard IoU metrics may underestimate the true segmentation quality of our best models.

\paragraph{Simulation Constraints:} While the virtual greenhouse accurately simulates spatial and structural properties, it currently ignores dynamic biomechanical interactions such as wind-induced foliage movement, which may limit its application in continuous video-tracking tasks \cite{Tang2020}.

\section{Conclusion}\label{sec6}
This work demonstrates that combining procedurally generated synthetic data with foundation model fine-tuning yields a robust pipeline for agricultural semantic segmentation. Our fully fine-tuned SAM~3 achieves the best overall real-domain performance across multiple external datasets, while weight interpolation offers complementary gains on individual datasets. Critically, our analysis reveals that synthetic-domain metrics alone are insufficient predictors of real-world generalization: the synthetic–real inversion observed across fine-tuning strategies shows that models ranking highest on synthetic validation do not necessarily transfer best to real imagery. This underscores that real-domain evaluation, rather than synthetic performance alone, is essential for reliable model selection and deployment. By releasing our fine-tuned SAM~3 weights, procedural model, and synthetic datasets, we provide the agricultural research community with an out-of-the-box tool for downstream phenotyping, accelerating the development of scalable agricultural vision systems without training custom models from scratch.

Future work will investigate the trade-offs in performance between semantic and instance-level tasks without sacrificing semantic accuracy: preliminary experiments indicate that specializing SAM~3's semantic head may trade off against its native detection behavior, but a rigorous evaluation under a matched protocol across all fine-tuning variants, including Full FT, remains ongoing. We also plan to expand the procedural model to other crops, such as cucumber or potato, incorporate diffusion models to further diversify training data, and explicitly track fruit ripening stages.

\section*{Acknowledgment}

\begin{sloppypar}
    We thank our anonymous commercial greenhouse partner for providing the data collection facilities. We also thank the creators of the TomatoMAP, LaboroTomato, Rob2Pheno, TomatoOD, AgRobTomato, and Tomato Plantfactory datasets for enabling our evaluation.
\end{sloppypar}

\bibliographystyle{unsrtnat}
\bibliography{references}  

\appendix

\section{L-system Model Details}\label{appendix:l_system_model}

\subsection{Biological Basis} \label{appendix:model_bio_basis}

The cherry tomato plant (\textit{Solanum lycopersicum} var.\ cerasiform) is an herbaceous species with a growth habit that is typically indeterminate, though cessation depends on the cultivar \cite{Went1944tomato, TorresQuezada2023tomato}. The plant follows a sympodial growth rule: the apical meristem terminates by producing a flower cluster after generating 2--3 leaves. Growth is then continued by the strongest axillary bud located immediately below the flower cluster \cite{Went1944tomato, TorresQuezada2023tomato}.

Initial development consists of 7 to 10 vegetative leaves before the first inflorescence appears. The plant produces compound leaves consisting of multiple leaflets attached to a rachis. Both the inflorescence and infructescence are also compound, terminating in a flower or fruit. In greenhouse environments, plants are managed via the \textit{leaning and lowering} technique: once the plant reaches approximately 2.5-3 m in height, the support string is released, and the stem is shifted horizontally along a wire, lowering the spent basal portion of the stem.

\subsection{Tomato Plant Modeling Literature}\label{appendix:tomato_plant_modeling}

Tomato plants are one of the most studied crop species in plant model research because of their economic importance and the amount of data available on their growth dynamics.
The models can be broadly divided into functional‑structural plant models (FSPMs), coupling architectural growth with physiological processes, and procedural geometric models, which emphasize realistic geometry and texture for visual simulation.

FSPMs for tomato plants, such as those by Sarlikioti et al. \cite{Sarlikioti2011hpa}, Chen et al. \cite{Chen2014qea}, de Visser et al. \cite{DeVisser2014oii}, and Zhang et al. \cite{Zhang2020hrs}, have historically focused on plant–light interactions and crop performance within greenhouse environments. Vermeiren et al. \cite{Vermeiren2019qti} extended this work by investigating the impact of leaf morphology on simulation accuracy. More recently, Butturini et al. \cite{Butturini2025afs} developed a dwarf-tomato FSPM optimized for vertical farming to analyze planting density at both the plant and canopy scales. Smolenova et al. \cite{Smolenova2025dtf} introduced a digital twin framework that couples a conventional FSPM with high-throughput phenotyping data, allowing real-time comparisons between simulated and measured crop states. Although FSPMs offer deep insight into plant–environment interactions, they often omit the photorealistic rendering required to generate high-quality synthetic datasets for machine-learning pipelines.

Conversely, procedural geometric models prioritize visual appearance but often do not simulate complex agronomic practices. While these type of 3D models are available from specialized software like Xfrog \cite{Lintermann1999Xfrog} and provide high-fidelity snapshots, they are not tailored for simulating life-cycle dynamics or phenotypic variation. L-systems bridge this gap by combining rule-based development with structural repetition. Cieslak et al. \cite{Cieslak2021lmi} proposed an interactive method to develop and visually calibrate L-system models, and Khan et al. \cite{Khan2025etp} demonstrated its utility in generating synthetic data for crops like corn and canola. Alternative frameworks, such as Blender’s Geometry Nodes, have also been employed. Klein et al. \cite{Klein2024sds} used procedural methods to generate geometry and textures for individual branches of tomato plants.

Other work has focused on specific physical interactions. Deng et al. \cite{Deng2024gps} developed a physically-based tomato model within the Gazebo environment, prioritizing kinematic constraints for robotic manipulation over growth dynamics. Maggioli et al. \cite{Filippo2023api} simulated internal water transport to model the wilting process, producing realistic deformations through physics-based solvers. Additionally, Wang et al. \cite{Wang2022phe} utilized Structure-from-Motion (SfM) on multi-view RGB images to reconstruct tomato growth and extract phenotypic traits from 3D meshes.

Building on the architectural framework established by Smolenova et al. \cite{Smolenova2025dtf}, we propose an L-system-based model that preserves the fundamental developmental rules of the tomato plant while abstracting away intensive functional simulations. This simplification enables the rapid generation of diverse synthetic imagery, including a simulation of greenhouse leaning and lowering techniques.

\subsection{Model Construction}

The model was developed and manually calibrated using an interactive method based on L-systems \cite{Cieslak2021lmi}. L-systems reduce structural complexity by exploiting the inherent modular repetition and self-similarity in plant development \cite{Prusinkiewicz1990tab}. The model operates on discrete modules: apices ($A$), internodes ($I$), leaves ($L$), and flowers/fruit ($K$), that are subject to three types of rules:
\begin{itemize}
    \item Production Rules: Ordinary parametric L-system rules advance the state of modules over time.
    \item Decomposition Rules: Define the hierarchy of compound modules; for example, a metamer is decomposed into an internode, a leaf, and an axillary bud.
    \item Interpretation Rules: Map the L-system string to 3D space using turtle geometry, defining the visual and geometric attributes of each module.
\end{itemize}
The simulation focuses on vegetative and reproductive phases, initiated by a primary apex. Due to the sympodial nature of the tomato, the model utilizes the same growth functions and target sizes for successive axes, as the growth pattern repeats once the axillary bud takes over.

\subsection{Implementation in L+C}

The model is implemented in the L+C modeling language \cite{Prusinkiewicz2007tlc}, which integrates L-system structures within a C++ context for high-performance execution.
Each module is defined by a set of parameters, including age $a$, metamer number $n$ along the parent axis, branching order $o$, and state $s$ (vegetative, reproductive, or inactive).
All components age by a fixed increment $dt$; for example, the rule for an apex is:
$$A(a, n, o, s) \rightarrow A(a + dt, n, o, s)$$
The apices also follow a decomposition rule producing a metamer $M$.
Depending on the state $s$, $M$ produces either vegetative components ($I, L, A$) or reproductive components ($K$).
An apex produces a predefined number of vegetative metamers before switching to a reproductive state to produce a flower cluster, after which it becomes inactive.
The plastochron (the interval between metamer production) is assumed to be constant.
Spiral phyllotaxis is simulated by a divergence angle of $137.5^\circ$ between successive metamers.

\subsection{Geometry and Visual Representation}

Internodes are modelled as cylinders with length and radius governed by graphical sigmoid functions of age and their position on the parent axis \cite{Prusinkiewicz2001tuo}. Compound leaves consist of 6--8 leaflets, alternating between large primary and small secondary leaflets. The model utilizes alpha-mapped textures to generate high-quality meshes based on selected tomato leaflet images from the PlantVillage dataset \cite{Mohanty2016udl}. Leaflets are assigned random meshes from a collection of five primary and five secondary meshes, with additional randomization applied to colour and brightness.

\begin{figure}[ht]
    \centering
    \includegraphics[width=0.75\textwidth]{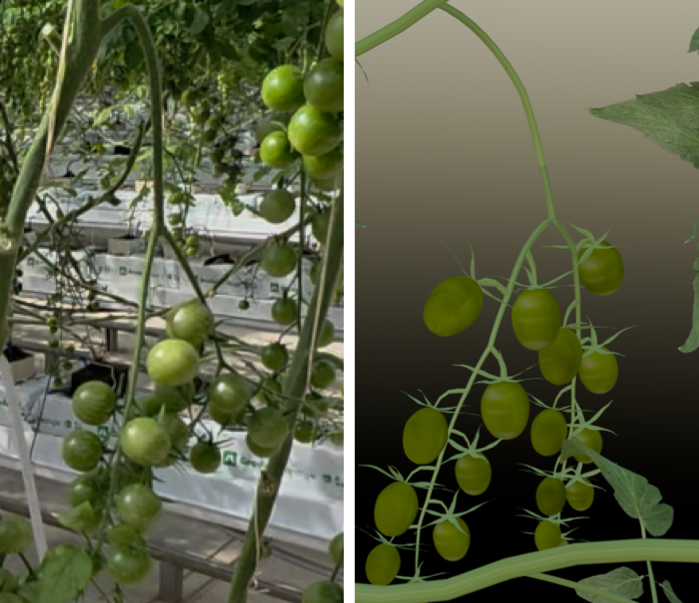}
    \caption{Simulation of bifurcated infructescence: photograph (left) vs. model visualization (right).}\label{fig:model-fruit}
\end{figure}

In variation from the typical single-branch structure of a tomato inflorescences, the main peduncle was modelled to bifurcate into two distinct branches, with each bearing flowers (see Fig.~\ref{fig:model-fruit}).
Flowers are modelled as five sepals and petals in a radially symmetric pattern around central stamens. Flower opening is a function of age, affecting the size and angle of the organs. Upon reaching a maximum age, the flower module is replaced by a fruit module, which is visualized as a surface of revolution based on a reference image.

\subsection{Tropism and Biomechanical Simulation}

To capture the natural posture of the plant, the model incorporates phototropism and gravitropism using L+C’s built-in tropism simulation \cite{Prusinkiewicz1990tab,Prusinkiewicz2007tlc}. The bending response of each module is governed by an elasticity parameter ($e$), which determines the degree to which the turtle's heading $\vec{H}$ aligns with the tropism vector $\vec{T}$.
This parameter is modelled as a function of the organ's age and type:
\begin{itemize}
    \item For stems and inflorescences, the gravitropism ($e_g$) is varied as fruit clusters age and increase in mass, causing the clusters to sag.
    \item For leaves, the model combines phototropism ($e_p$) to orient the lamina toward light and gravitropism to account for the weight of the compound structure.
\end{itemize}
In the current iteration of the model, collision detection and resolution are ignored because avoiding self-intersections is computationally expensive for individual plants and for collections of plants. Branching and bending are driven strictly by the interaction of growth rules and tropism vectors; consequently, physical intersections between plant components may occur.

\subsection{Simulating Greenhouse Management}

The simulation environment is developed using Unreal Engine~5 (UE5), which provides photorealistic rendering capabilities and physically-based scene representation suitable for high-fidelity synthetic data generation.

To enable sensor simulation and programmatic data acquisition, we integrate Cosys-AirSim~\cite{jansen2023cosys}, a maintained extension of the original AirSim framework offering multi-sensor simulation and a Python API interface. Built on this foundation, the simulation pipeline comprises two main modules: a procedural greenhouse generator created with the Unreal Engine~5 Python interface and a data collection system driven by the Cosys-AirSim API. Together, these modules enable the creation of highly configurable, physics-accurate greenhouse environments for generating synthetic data.

The greenhouse generation pipeline offers extensive flexibility, allowing the reproduction of a wide range of real-world greenhouse configurations. Key structural parameters are explicitly defined, including the number of planting rows and columns, inter-row spacing, intra-row plant spacing, plant distribution patterns, and ceiling height. Based on these parameters, the overall greenhouse dimensions are automatically adjusted to ensure geometric coherence and spatial consistency. This design enables rapid adaptation of the environment to different crop types, planting densities, and spatial layouts.
To further enhance realism and dataset diversity, plant growth stage assignment can be configured as either deterministic or randomized. This functionality enables the simulation of heterogeneous growth conditions within the same greenhouse, reflecting natural variations encountered in operational settings and thereby improving the generalization of downstream learning models. In our experiments, the greenhouse configuration follows standard practices in tomato cultivation, reproducing typical spacing between plants, columns, and elevation.

\begin{figure}[ht]
    \centering
    \includegraphics[width=0.75\textwidth]{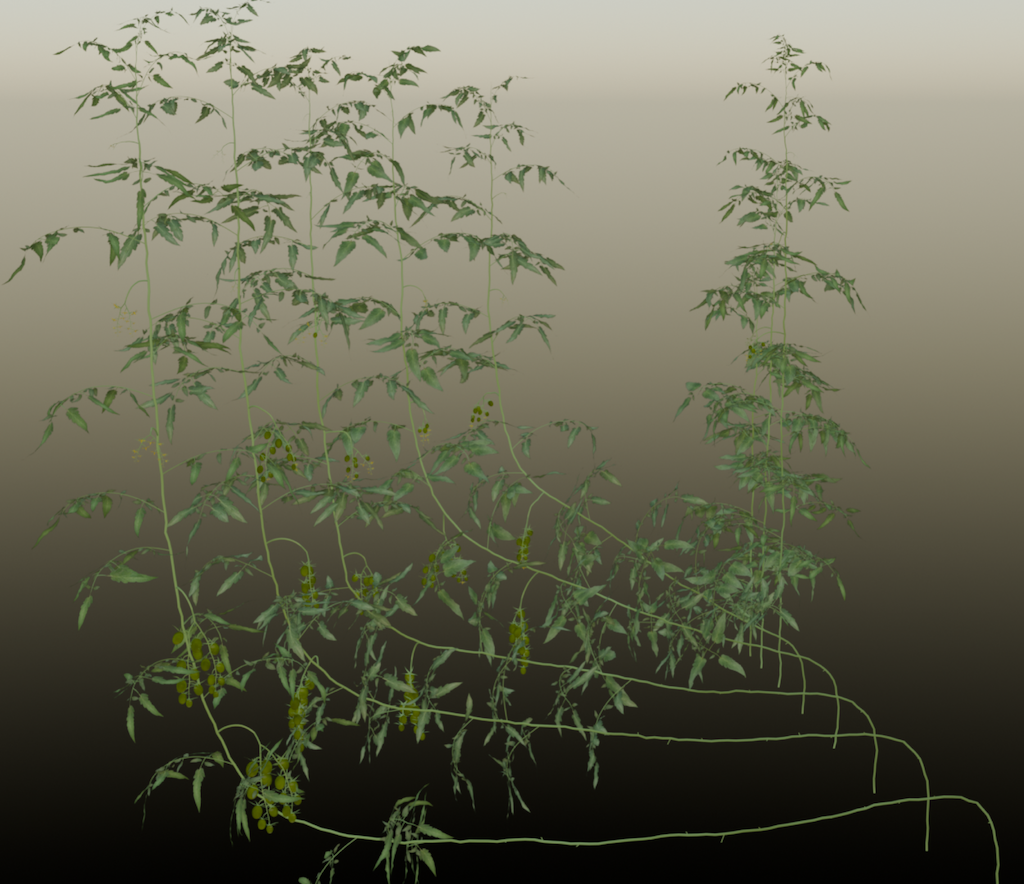}
    \caption{Simulation of the leaning and lowering management practice.}\label{fig:model-lowering-leaning}
\end{figure}

The interaction between the plant and its support wire was implemented using Environmental L-systems \cite{Mech1996vmp}. The \texttt{GetPos()} module was used to query the current spatial coordinates of the apex. When the apical height exceeded 2.5 m, the model dynamically adjusted the turtle’s heading by interpolating its direction toward a user-defined curve, simulating the leaning and lowering management practice (see Fig.~\ref{fig:model-lowering-leaning}). The stem geometry is modified by interpolating the turtle’s heading direction between two states: (1) a standard vertical function ($y = 0$), and (2) a graphically defined function tracing the horizontal/lowered shape of the stem. As the plant is lowered, the model also simulates pruning by dropping the oldest leaves and spent fruit clusters based on their age.

By adjusting growth parameters within the generative model, plants are synthesized at different phenological stages, spanning early vegetative to mature fruit-bearing phases. The models are further processed to separate semantic components (leaves, stems, flowers, and fruits) via material decomposition, enabling controlled manipulation of plant morphology, fruit distribution, and canopy structure for structured synthetic data generation.

To bridge the procedural generation and photo-realistic rendering phases, we developed an automated ingestion and annotation pipeline connecting the L-system geometric output to Unreal Engine~5. This pipeline handles geometric parsing, dynamic shading, and automated ground-truth extraction. The L-system-based plant simulator exports the generated plant structures as Wavefront .obj files, encoding both the geometry and hierarchical organization.
It was used to generate approximately 50 unique plant assets, which were then ingested into Unreal Engine~5 via an automated Python API script. Each plant was instantiated and each semantic component (leaves, stems, flowers, and fruits) was assigned a distinct Material Slot based on its upstream L-system module ID.

\section{Synthetic Dataset Acquisition Details}\label{appendix:synthetic_dataset}

The model is capable of producing hundreds of unique plant instances at diverse developmental stages. On this basis, a high-variance synthetic dataset is designed for greenhouse scene reconstruction and for training machine learning algorithms, where structural diversity is essential for robust model generalization. The framework provides a controlled yet flexible foundation for generating greenhouse scenes and training computer vision models for tasks such as organ detection and yield estimation.

\begin{table}[ht]
    \centering
    \caption{Generated dataset structure: Content associated to each image}
    \footnotesize
    \renewcommand{\arraystretch}{1.1}
    \begin{tabular}{|c|c|}
        \hline
        \textbf{Features}   & \textbf{Description}                                \\ \hline
        Timestamp           & Acquisition time                                    \\ \hline
        Image ID            & Index of the image taken                            \\ \hline
        Camera Settings     & Field of view (FOV) and resolution                  \\ \hline
        RGB Image           & Red, Green, Blue image                              \\ \hline
        Depth Image         & Distance of each pixel from the camera              \\ \hline
        Bounding Box Center & Center coordinates of the bounding box              \\ \hline
        Bounding Box Extent & Half-dimensions of the bounding box                 \\ \hline
        \hline
        \textbf{Labels}     & \textbf{Description}                                \\ \hline
        Semantic Image      & Pixel label (leaf, stem, flower, fruit, background) \\ \hline
        Growth Stage        & Indicates growth stage or background                \\ \hline
        Instance Image      & Identifies and differentiates plants                \\ \hline
    \end{tabular}
    \label{tab:dataset_structure_img}
\end{table}

To systematically capture the simulated scenes, we implement a configurable image acquisition pipeline supporting multiple virtual camera placements within the greenhouse environment. Camera position and orientation are parameterized to emulate realistic observation viewpoints, including lateral inter-row perspectives and elevated overview views. For each acquisition step, structured metadata are recorded, including timestamp, camera position $(x, y, z)$, and intrinsic parameters such as field of view (FOV) and image resolution.

The content of the dataset is described in Table~\ref{tab:dataset_structure_img}. At every capture instance, the simulator generates synchronized RGB images and depth maps, where each pixel in the depth image encodes its distance from the camera. In parallel, annotations are automatically derived from the simulation engine. These include 2D bounding box information for each plant instance (center coordinates and spatial extent), pixel-wise semantic labels (leaf, stem, flower, fruit, background), plant growth stage, and instance identifiers. Since plant geometry and spatial configuration are explicitly defined within the environment, all labels maintain precise correspondence with the rendered images.

This acquisition framework results in a fully labeled synthetic dataset tailored for 2D computer vision applications. We emphasize the ``multi-modal completeness'' of these pixel-perfect annotations (Depth + Instance + Semantic) as a benchmark-ready package for the community.
It supports instance segmentation to separate individual plants in dense greenhouse rows and semantic segmentation for organ-level classification. The structured metadata and hierarchical labeling scheme provide a comprehensive and consistent basis for training and evaluating learning-based models under controlled greenhouse conditions.

\subsection{Stochastic Variation and Dataset Generation}

\begin{figure}[htp]
    \centering
    \includegraphics[width=0.5\textwidth]{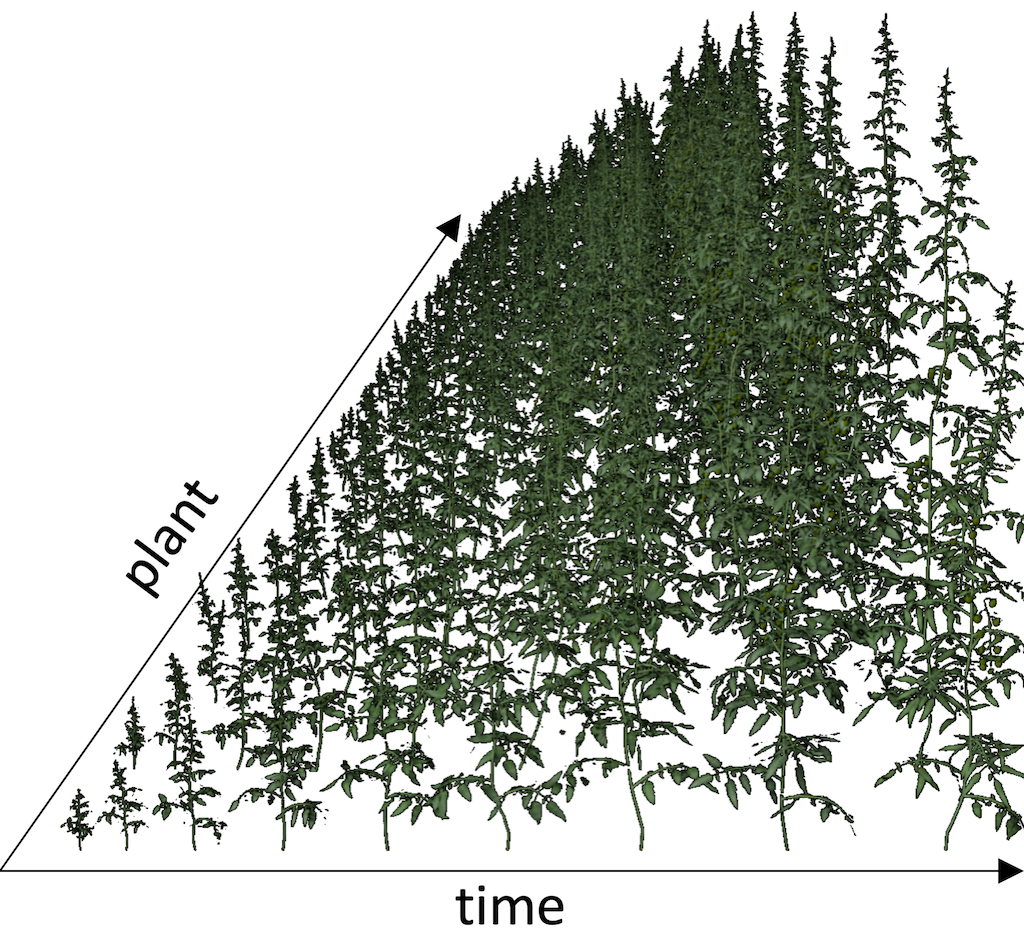}
    \caption{Variation in plant morphology over time.}\label{fig:model-variation}
\end{figure}

To account for natural biological variation and to facilitate the generation of large-scale datasets, a stochastic component was introduced into the model \cite{Cieslak2021lmi}. Key geometric parameters, such as the scale and orientation of leaves, flowers, and fruit, were modulated by a multiplier $r$:
$$r \sim \mathcal{N}(\mu, \sigma^2)$$
where the mean $\mu = 1.0$ and the standard deviation $\sigma \leq 0.1$.
The model further randomizes the divergence angle of the spiral phyllotaxis (centered at $137.5^\circ$) and the branching angles of both leaves and inflorescences.
This variation ensures that no two plants are morphologically identical, despite sharing the same underlying production rules. By varying the random seed and the simulation time (the total number of derivation steps), the model was used to generate hundreds of unique plant instances at different developmental stages. Figure \ref{fig:model-variation} shows some of these plants at different development stages before leaning and lowering is applied.

\end{document}